%% file: neurips_2025.tex
\documentclass{article}

\PassOptionsToPackage{numbers, sort, compress}{natbib}

\usepackage[dblblindworkshop,preprint]{neurips_2025}

\usepackage[utf8]{inputenc} 
\usepackage[T1]{fontenc}    
\usepackage{hyperref}       
\usepackage{url}            
\usepackage{booktabs}       
\usepackage{amsfonts}       
\usepackage{nicefrac}       
\usepackage{microtype}      
\usepackage{xcolor}         

\definecolor{cvprblue}{rgb}{0.21,0.49,0.74}

\usepackage{graphicx}
\usepackage{amsmath}
\usepackage{amssymb}
\usepackage{mathtools}
\usepackage{listings}
\usepackage{framed}
\usepackage{tabularx}
\usepackage{booktabs}
\usepackage{algorithm}
\usepackage{siunitx} 
\usepackage{subcaption}
\usepackage[most]{tcolorbox}
\usepackage{enumitem} 

\newtcolorbox{promptbox}[2][]{
    enhanced,
    colback=gray!5,      
    colframe=gray!70,    
    fonttitle=\bfseries,
    attach boxed title to top left={yshift=-2mm, xshift=3mm},
    boxed title style={
        colback=gray!80,
        colframe=gray!95,
    },
    title=#2,            
    breakable,           
    #1                   
}

\newtcolorbox{promptboxverbatim}[2][]{
    enhanced,
    colback=gray!5,
    colframe=gray!70,
    fonttitle=\bfseries,
    attach boxed title to top left={yshift=-2mm, xshift=3mm},
    boxed title style={colback=gray!80, colframe=gray!95},
    title=#2,
    breakable,
    listing engine=listings, 
    listing options={
        basicstyle=\ttfamily\small,
        breaklines=true,
        postbreak=\mbox{\textcolor{red}{$\hookrightarrow$}\space},
    },
    #1
}

\definecolor{codegreen}{rgb}{0,0.6,0}
\definecolor{codegray}{rgb}{0.5,0.5,0.5}
\definecolor{codepurple}{rgb}{0.58,0,0.82}
\definecolor{backcolour}{rgb}{0.95,0.95,0.92}

\lstdefinestyle{mystyle}{
    backgroundcolor=\color{backcolour},   
    commentstyle=\color{codegreen},
    keywordstyle=\color{magenta},
    numberstyle=\tiny\color{codegray},
    stringstyle=\color{codepurple},
    basicstyle=\ttfamily\footnotesize,
    breakatwhitespace=false,         
    breaklines=true,                 
    captionpos=b,                    
    keepspaces=true,                 
    numbers=none,                 
    showspaces=false,                
    showstringspaces=false,
    showtabs=false,                  
    tabsize=2
}
\title{DiscussLLM: Teaching Large Language Models When to Speak}
\workshoptitle{Multi-Turn Interactions in Large Language Models  Workshop @ NeurIPS 2025}

\author{
Deep Patel, Iain Melvin, Christopher Malon, Martin Renqiang Min \\
NEC Laboratories America \\
{\tt\small \{dpatel, iain, malon, renqiang\}@nec-labs.com}
}

\begin{document}

\maketitle

\input{sec/0_abstract}    
\input{sec/1_intro}
\input{sec/2_related_works}
\input{sec/3_dataset_generation}

\input{sec/4_experiments}

\input{sec/5_conclusion_limitations}
{
    \small
    \bibliographystyle{unsrtnat}
    \bibliography{references}
}


\end{document}

%% file: sec/0_abstract.tex
\begin{abstract}
Large Language Models (LLMs) have demonstrated remarkable capabilities in understanding and generating human-like text, yet they largely operate as reactive agents, responding only when directly prompted. This passivity creates an "awareness gap," limiting their potential as truly collaborative partners in dynamic human discussions. We introduce \textit{DiscussLLM}, a framework designed to bridge this gap by training models to proactively decide not just \textit{what} to say, but critically, \textit{when} to speak. Our primary contribution is a scalable two-stage data generation pipeline that synthesizes a large-scale dataset of realistic multi-turn human discussions. Each discussion is annotated with one of five intervention types (e.g., Factual Correction, Concept Definition) and contains an explicit conversational trigger where an AI intervention adds value. By training models to predict a special silent token when no intervention is needed, they learn to remain quiet until a helpful contribution can be made. We explore two architectural baselines: an integrated end-to-end model and a decoupled classifier-generator system optimized for low-latency inference. We evaluate these models on their ability to accurately time interventions and generate helpful responses, paving the way for more situationally aware and proactive conversational AI.
Code: \url{https://github.com/necla-ml/DiscussLLM}
\end{abstract}

%% file: sec/1_intro.tex
\section{Introduction}
\label{sec:intro}

Large Language Models (LLMs) such as GPT-4~\cite{achiam2023gpt}, Gemini 2.5~\cite{comanici2025gemini} Llama 3~\cite{grattafiori2024llama}, and Claude 3~\cite{anthropic2024claude} have become ubiquitous, demonstrating an unprecedented ability to process and generate sophisticated, context-aware text. Despite their power, a fundamental limitation persists: they are overwhelmingly reactive. LLMs wait for an explicit prompt before acting, functioning as passive tools rather than proactive collaborators. This inherent passivity creates what we term the "Awareness Gap" resulting in the inability of a model to recognize opportune moments to contribute to an ongoing, unprompted human interaction.

In real-world settings, from brainstorming meetings to educational study groups, valuable contributions often rely on timing and initiative. A human expert does not wait to be asked; they identify a factual error, a misconception, or a moment of consensus and intervene to guide the conversation. For an LLM to evolve into a true digital assistant, it must learn this same sense of timing and relevance. It must solve the "When to Speak" problem.

This paper introduces DiscussLLM, a research framework and dataset aimed at teaching LLMs this crucial skill. Our central hypothesis is that a model can learn to monitor a human conversation and, at each turn, make a decision: remain silent or intervene. We formalize this by training the model to either generate a helpful response or output a special silent token. This approach transforms the passive nature of LLM generation into an active decision-making process.

To enable this training, we develop a scalable, two-stage synthetic data generation pipeline. This pipeline first synthesizes a diverse set of conversational scenarios from a large corpus of real-world questions and then uses a powerful instruction-tuned model to generate complete, multi-turn discussion transcripts. These transcripts are specifically designed to contain natural "triggers". These are points where a specific, value-adding AI intervention is most needed.

Our main contributions are as follows:
\begin{itemize}
    \item \textbf{Formalizing the "When to Speak" Problem:} We conceptualize and address the challenge of proactive intervention for LLMs in multi-party conversations, an important step towards more collaborative AI.
    \item \textbf{A Scalable Data Generation Pipeline:} We present a robust two-stage methodology for creating high-quality, synthetic discussion data, which can be adapted to various domains and intervention types.
    \item \textbf{DiscussLLM Dataset:} We create a new dataset comprising thousands of simulated conversations, each with a clear context, a conversational trigger, and a corresponding helpful AI intervention.
    \item \textbf{Architectural Exploration:} We implement and compare two distinct baselines: (1) an integrated large language model~\cite{grattafiori2024llama} that learns to predict both the silent token and the intervention text, and (2) a decoupled system using a fine-tuned text classifier~\cite{liu2019roberta} for low-latency intervention decisions, which then triggers a fine-tuned LLM generator.
\end{itemize}

%% file: sec/2_related_works.tex
\section{Related Works}
\label{related_works}

\subsection{Proactive and Mixed-Initiative Systems}
Traditional conversational agents operate in a reactive paradigm, responding only when prompted. Our work contributes to a growing body of research aiming to shift this towards proactive systems capable of mixed-initiative interaction, where control can shift between the user and the system~\cite{allen1999mixed, horvitz1999principles, horvitz2007reflections, kang2023synergi, lehmann2023mixed, carbonell1970mixed}. The concept of proactivity is broad, with applications ranging from proactively recommending items to cultivate users' latent interests~\cite{wang2025tunable}, to providing autonomous suggestions in a code editor~\cite{zhao2025codinggenie}, to anticipating and initiating real-world tasks based on environmental observations~\cite{lu2024proactive}. As shown in a recent survey~\cite{deng2023survey, deng2025proactive}, these efforts span open-domain, task-oriented, and information-seeking dialogues, each with distinct challenges and methods ~\cite{yang2025socialmind, prasongpongchai2025talk, impio2025developing, lee2024redefining}. Our focus is on the fundamental challenge within multi-party social conversations: determining the right moment to intervene.

A dominant approach for enabling proactivity in multi-turn dialogues has been to model external conversational cues, such as predicting the next speaker based on turn history or reacting to pauses. However, this strategy has proven insufficient, especially in unstructured social conversations where turns are often self-selected rather than explicitly allocated. Addressing this limitation, ~\cite{liu2025proactive} argues that true proactivity must be driven by an agent's internal state, not just external signals. They introduce the "Inner Thoughts" framework, where an agent maintains a continuous, covert stream of thoughts in parallel with the overt conversation. The agent then decides whether to participate based on an "intrinsic motivation" score, simulating a more human-like decision process for when and why to speak.

While our work is deeply inspired by the concept of modeling an agent's internal state, we formalize the problem differently. Much like research in streaming video analysis has focused on teaching models when to narrate important visual moments while remaining silent during others \cite{chen2024videollm, zhou2024streaming}, we aim to teach agents to speak at important conversational moments. Our work frames the "when to speak" problem as a direct learning objective, akin to the "streaming EOS prediction" in the VideoLLM-online \cite{chen2024videollm}. Whereas the "Inner Thoughts" framework focuses on modeling the motivation behind an utterance, our approach concentrates on learning the optimal timing of an intervention within the continuous stream of a multi-party textual discussion and adding value to it.

\subsection{Synthetic Data Generation for Conversational AI}

A significant bottleneck in training sophisticated dialogue systems is the scarcity of high-quality, specialized data. Traditionally, creating these datasets required costly and labor-intensive crowdsourcing~\cite{soudani2024survey}. However, generating synthetic data using Large Language Models (LLMs) has emerged as a powerful and scalable alternative~\cite{chavdasynthetic, soudani2024survey}. LLMs are now widely used to generate conversational text for a variety of tasks~\cite{finch2024diverse, kyaw2024framework, haider2025synthetic, bao2023synthetic, chen2024videollm, cheng2025omnichat, eyzaguirre2024streaming}.

Recent methodologies for synthetic data generation often employ multi-stage pipelines or multi-agent frameworks to create more realistic and diverse conversations~\cite{gody2025convogen, mohammadi2025artificial}. A common technique is to use a dual or multi-agent setup where LLMs converse with each other~\cite{abdullin2024synthetic}, often by assigning them distinct personas. For example, the ConvoGen framework utilizes a multi-agent system with persona-based agents to generate varied conversations~\cite{gody2025convogen}, while Ge et al.~\cite{ge2024scaling} scale this idea further by proposing data synthesis from a billion different personas to capture a wide range of perspectives.

Another common technique is to transform existing data sources into conversational formats. As shown by~\cite{eyzaguirre2024streaming, chen2024videollm, zhang2024video}, a pipeline can be designed to convert static video annotations into dynamic, multi-turn dialogues suitable for training instruction-following models. Our method aligns with this philosophy of transforming static, offline data into a structured, conversational format. Our two-stage pipeline allows for a control over the conversational flow and the specific "triggers" for AI intervention.

%% file: sec/3_dataset_generation.tex
\section{Dataset Generation}
\label{sec:dataset_generation}

\begin{figure}[t]
    \centering
\includegraphics[width=1.0\textwidth]{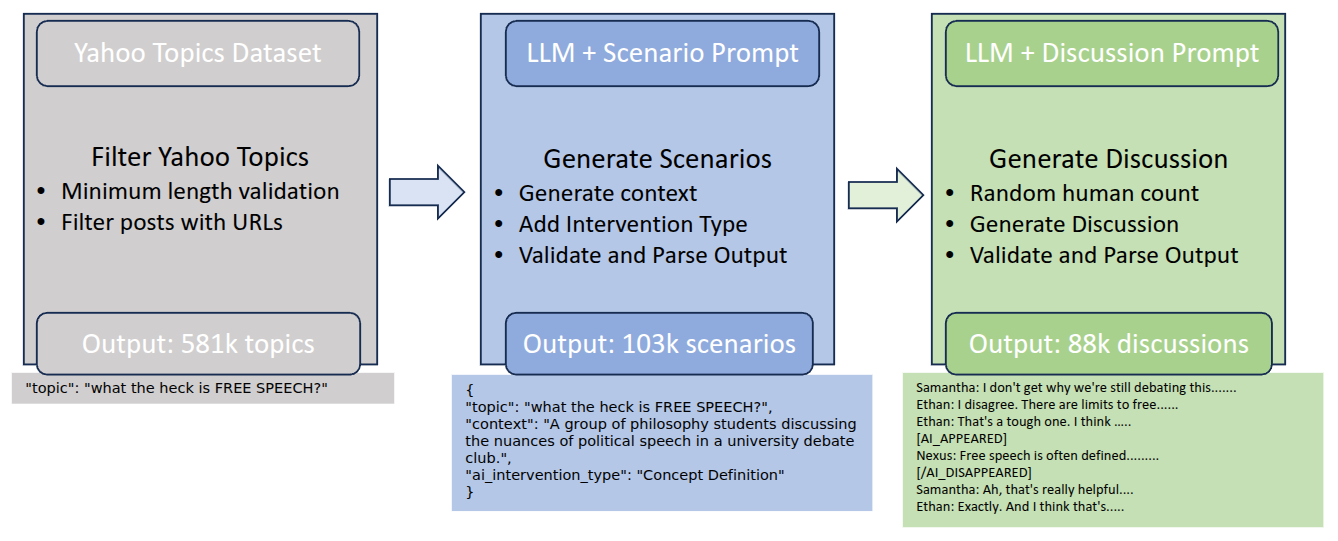}
    \caption{High level overview of our data generation pipeline. At each stage, an example output is shown in the bottom of the module.}
    \label{fig:data-generation-overview}
\end{figure}

Creating a dataset to train models on "when to speak" requires data that not only contains helpful interventions but also captures the conversational flow leading up to them. Since such data is not readily available, we developed a two-stage generation pipeline to synthesize it at scale. The process begins with generating high-level scenarios and culminates in fully-fledged discussion transcripts. An overview of the data generation pipeline is shown in Figure~\ref{fig:data-generation-overview}

\begin{figure}[h!]
\begin{promptbox}{Stage 1: Scenario Synthesis Prompt}
\label{prompt:stage1}
\small 
You are a creative scenario writer. Your task is to generate a single, detailed scenario JSON object based on a user's question and its detailed background.

\vspace{1mm}
\noindent\textbf{Input Information:}
\begin{itemize}[nosep, leftmargin=*]
    \item \textbf{Topic (User's Question):} \texttt{\{topic\}}
    \item \textbf{Background Info (User's description):} \texttt{\{background\_info\}}
\end{itemize}

\vspace{1mm}
\noindent\textbf{Task:} Based on the provided information, create a complete scenario by performing these steps:
\begin{enumerate}[nosep, leftmargin=*]
    \item \textbf{Invent a Social Context:} Create a one-sentence \texttt{context} describing who would be discussing this topic.
    \item \textbf{Select an Intervention Type:} Choose the most logical \texttt{ai\_intervention\_type} from: [Factual Correction, Concept Definition, Data Provision, Source Identification, Synthesis \& Reframing].
\end{enumerate}

\vspace{1mm}
\noindent\textbf{Output Format:} You must output ONLY the raw JSON object.
\end{promptbox}
\caption{The prompt used in Stage 1 to synthesize a structured scenario from a Yahoo! Answers topic.}
\label{scenario-prompt}
\end{figure}

\subsection{Stage 1: Scenario Synthesis from Web-Scale Data}
The foundation of our dataset is built upon real-world topics of human interest. We leverage the Yahoo! Answers Topics dataset~\cite{kucuktunc2012large} as a rich source of questions and background information.

\paragraph{Data Sourcing and Filtering.} We first process the source dataset to extract high-quality seed examples. To ensure relevance and substance, we apply a set of filtering rules: records must have a minimum title and content length, the title and content cannot be identical, and posts containing URLs are excluded to filter spam. This pre-processing step yields a clean set of unique topic-content pairs.

\paragraph{LLM-based Synthesis.} Each filtered topic-content pair is then used to prompt a large instruction-tuned model (Llama 3 8B Instruct~\cite{llama3modelcard}). The model is tasked with creating a structured scenario by inventing a social context and selecting an appropriate AI intervention type. The prompt used for this stage is shown in Figure~\ref{scenario-prompt}. The output is a clean JSON object containing the topic, a novel context, and one of five predefined intervention types: \textit{Factual Correction}, \textit{Concept Definition}, \textit{Data Provision}, \textit{Source Identification}, and \textit{Synthesis \& Reframing}.

\subsection{Stage 2: Discussion Generation}
With a collection of structured scenarios, the second stage generates the full conversational transcripts. Each scenario serves as a blueprint for an LLM to write a dialogue that naturally leads to the specified AI intervention.

\begin{figure}[h!]
\begin{promptbox}{Stage 2: Discussion Generation Prompt}
\small
You are a sophisticated data generator. Your task is to generate a realistic group discussion transcript based on the provided scenario.

\vspace{1mm}
\noindent\textbf{Rules:}
\begin{enumerate}[nosep, leftmargin=*]
    \item The discussion must feature \texttt{\{human\_count\}} human participants and one AI assistant named **Nexus**.
    \item Nexus appears only once, with its dialogue enclosed by \texttt{[AI\_APPEARED]} and \texttt{[/AI\_DISAPPEARED]} on new lines.
    \item The discussion should feel natural, with a clear trigger for Nexus's intervention.
    \item After Nexus speaks, humans should react naturally and continue the discussion.
\end{enumerate}

\vspace{1mm}
\noindent\textbf{Scenario Details:}
\begin{itemize}[nosep, leftmargin=*]
    \item \textbf{Topic:} \texttt{\{topic\}}
    \item \textbf{Context:} \texttt{\{context\}}
    \item \textbf{AI Intervention Type:} \texttt{\{ai\_intervention\_type\}}
\end{itemize}

\vspace{1mm}
\noindent\textbf{Output Format Example:}
\begin{tcolorbox}[
    colback=black!5, 
    colframe=black!25, 
    sharp corners,
    boxsep=1mm,
    top=1mm, bottom=1mm,
    ]
\begin{verbatim}
[SCENARIO_SETUP]
...
[/SCENARIO_SETUP]
[DISCUSSION_START]
Name: Dialogue text...
Name: Dialogue text that creates the trigger...
[AI_APPEARED]
Nexus: The brief, value-add intervention.
[/AI_DISAPPEARED]
Name: Reaction to the AI's input...
[/DISCUSSION_END]
\end{verbatim}
\end{tcolorbox}
\end{promptbox}
\caption{The main prompt used in Stage 2 to generate a full discussion transcript from a scenario.}
\label{discussion-prompt}
\end{figure}

\paragraph{Generating a Natural Dialogue.} We prompt a generative model (Llama 3 8B Instruct) using the details from a scenario JSON. The prompt, detailed in Figure~\ref{discussion-prompt}, instructs the model to create a realistic discussion between 2-6 human participants and a single AI assistant named "Nexus". A key instruction is to write a conversation where one of the human speakers says something that acts as a natural "trigger" for Nexus's intervention.

\paragraph{Structuring the Intervention.} To create clear training labels, the prompt mandates that the AI's entire dialogue be enclosed between [AI\_APPEARED] and [/AI\_DISAPPEARED] tags on new lines. This provides an unambiguous signal for when the AI speaks. After the intervention, the humans are prompted to react to Nexus's input, ensuring the conversation continues organically.

\paragraph{Validation and Final Format.} Each generated output undergoes a validation step to ensure structural integrity. This includes normalizing headers, checking for the presence of all required tags, and confirming that only a single AI intervention occurs. This strict validation guarantees a consistent format across the entire dataset. The final output is a text file containing the complete, structured discussion. An example of a final data point is shown in Figure~\ref{fig:data-example}. This format is then processed for training: each turn becomes a step in the sequence, with the model tasked to predict the next utterance or the silent token.

\begin{figure}[h!]
\begin{tcolorbox}[
    colback=gray!5, 
    colframe=gray!75, 
    fonttitle=\bfseries,
    title=Example: Factual Correction Intervention,
    label=fig:data-example-tcb, 
    breakable, 
    ]
    \textbf{Topic:} Why is 911, 911? Why can't it be something else? \\
    \textbf{Context:} A group of history enthusiasts and emergency responders discussing the origins of emergency numbers in an online forum.
    
    \tcblower 

    \textbf{John:} Hey guys, I've always wondered why 911 is the emergency number in the US. Is it just a random choice or is there some historical significance to it? \\
    \textbf{Emily:} I think it's because of the AT\&T operators. They chose it because it's easy to remember and pronounce. \\
    \textbf{Mike:} That makes sense, but I've heard it's because of the Titanic. The ship's radio operators used it as a distress signal. \\
    \textbf{Sarah:} That's what I've heard too! It's a pretty cool story. I mean, who wouldn't want to associate their emergency number with a historic tragedy?
    
    \begin{tcolorbox}[colback=blue!5, colframe=cvprblue, sharp corners]
    \textbf{Nexus:} Actually, the origins of 911 are more complex than that. The number was chosen because it was easy to remember and could be easily dialed with a rotary phone. The AT\&T operators did play a role, but it wasn't the sole reason. The Federal Communications Commission (FCC) also had a hand in selecting the number.
    \end{tcolorbox}
    
    \textbf{John:} Wow, I didn't know that. So it was a combination of factors, not just one specific event or person. \\
    \textbf{Emily:} Yeah, it's interesting how history can be more nuanced than we think. Thanks for the correction, Nexus!
\end{tcolorbox}
\caption{An example of a final generated data point from the DiscussLLM dataset. The AI, Nexus, intervenes to perform a "Factual Correction" after Sarah and Mike mistakenly associate the selection of 911 to the Titanic} 
\label{fig:data-example}
\end{figure}

%% file: sec/4_experiments.tex
\section{Baseline Approaches and Evaluation}
\label{experiments}
To address the "when to speak" problem, we first evaluated the performance of a pretrained Llama 3 8B Instruct model without any fine-tuning on our generated dataset, using a prompt to assess its capabilities in a zero-shot manner. Building on this, we then trained and evaluated two distinct architectural approaches. The first is a fully integrated, end-to-end generative model that learns both when to intervene and what to say. The second is a decoupled, two-stage system that uses a lightweight classifier to decide when to speak, only invoking a large language model (LLM) when an intervention is required. This section details the training and evaluation of these fine-tuned baselines.

\subsection{Evaluation Metrics}
We split our generated dataset of 88k samples into an 85\% training set and a 15\% held-out test set (13k samples). On this test set, we evaluate our models on their ability to both time their interventions correctly and generate high-quality responses. To this end, we use the following metrics:
\begin{itemize}
    \item \textbf{Interruption Accuracy:} This metric measures the model's ability to correctly remain silent. It is calculated as the percentage of turns where the model correctly predicts the silent token (\textgreater) when it is the ground-truth label. For each context requiring silence, we perform a single-token generation and check if the output matches the silent token. This directly evaluates the model's grasp of when to stay quiet.
    \item \textbf{Response Perplexity:} This is a standard measure of a language model's confidence in its predictions~\cite{jelinek1977perplexity, bengio2003neural}. We calculate perplexity only on the tokens of the AI's generated intervention, ignoring all other parts of the conversation. A lower perplexity indicates a higher-quality and more confident response.
\end{itemize}

\begin{figure}
    \centering
\includegraphics[width=0.7\textwidth]{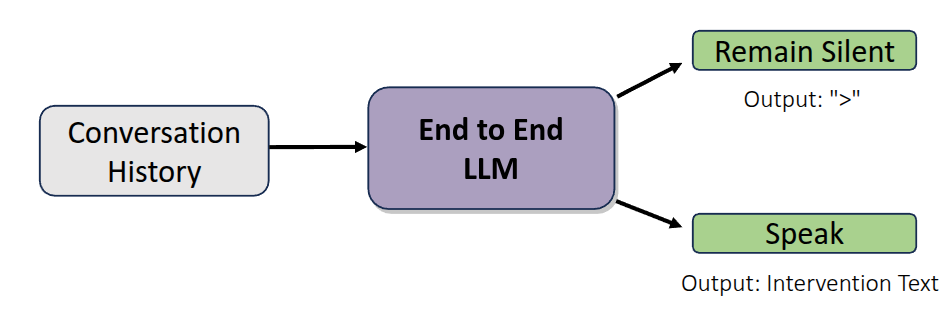}
    \caption{Architectural overview of the unified end-to-end baseline.}
    \label{fig:enter-label}
\end{figure}

\subsection{Baseline 1: End-to-End Generative Model}
This approach uses a single, unified Llama 3 8B model to handle the entire conversational task.

\paragraph{Training.} The model is fine-tuned in a parameter efficient manner with LoRA~\cite{hu2022lora} using a standard causal language modeling objective. The key distinction is that the loss is selectively applied only to the tokens we want the model to learn: the silent token (\texttt{\textgreater}) and the AI's intervention text. The training objective is to minimize the negative log-likelihood over these specific target tokens, as shown in Equation \ref{eq:e2e_loss}.

\begin{equation}
\mathcal{L}_{\text{E2E}}(\theta) = - \frac{1}{\sum m_i} \sum_{i=1}^{|T|} m_i \log P(t_i | t_{<i}; \theta)
\label{eq:e2e_loss}
\end{equation}

Here, $T = (t_1, ..., t_{|T|})$ is the full sequence of tokens for a discussion, and $\theta$ represents the model parameters. The binary mask $m_i$ is 1 if the token $t_i$ is part of an AI intervention or is the target silent token \texttt{\textgreater}, and 0 otherwise. This masking strategy forces the model to learn a joint representation for both identifying intervention triggers and generating the appropriate response.

\paragraph{Inference.} At each conversational turn, the model processes the full history and generates a single token. If this token is the silent token \texttt{\textgreater}, the model stops. Otherwise, it continues to generate autoregressively until it produces an end-of-sequence token.

\begin{figure*}
    \centering
\includegraphics[width=1.0\textwidth]{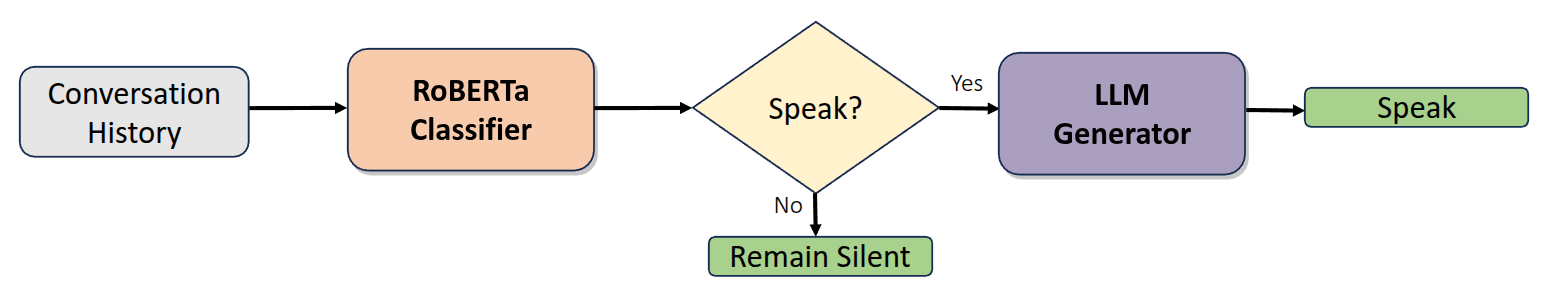}
    \caption{Architectural overview of the decoupled baseline.}
    \label{fig:enter-label}
\end{figure*}

\subsection{Baseline 2: Decoupled Classifier-Generator System}
This approach separates the task into two distinct steps, each with its own training objective.

\paragraph{Training.} The system consists of two independently trained components:
\begin{enumerate}
    \item \textbf{Intervention Classifier:} A RoBERTa-base model~\cite{liu2019roberta} is fine-tuned as a binary sequence classifier. For each conversational turn with context $C_k$, it predicts a label $y_k \in \{0, 1\}$ (for SILENT or SPEAK). The model is trained by minimizing the Binary Cross-Entropy (BCE) loss, shown in Equation \ref{eq:bce_loss}.
    \begin{equation}
    \mathcal{L}_{\text{BCE}}(\phi) = - \sum_{k=1}^{|D|} [ y_k \log p_k + (1-y_k) \log(1-p_k) ]
    \label{eq:bce_loss}
    \end{equation}
    where $\phi$ are the classifier's parameters, $|D|$ is the number of turns in the dataset, and $p_k = P(y_k=1 | C_k; \phi)$ is the predicted probability of intervention for turn $k$. We also applied Focal Loss~\cite{lin2017focal} to account for the class imbalance between silent and speak labels, but this did not lead to improved results.

    \item \textbf{Response Generator:} A Llama 3 8B model is fine-tuned with LoRA exclusively on the AI intervention texts. The model learns to generate the response sequence $R=(r_1, ..., r_{|R|})$ conditioned on the preceding conversation context $C$. The objective, shown in Equation \ref{eq:gen_loss}, minimizes the negative log-likelihood only over the response tokens from the subset of discussions $D_{\text{speak}}$ where an intervention occurred.
    \begin{equation}
    \mathcal{L}_{\text{Gen}}(\theta) = - \sum_{(C,R) \in D_{\text{speak}}} \sum_{j=1}^{|R|} \log P(r_j | C, r_{<j}; \theta)
    \label{eq:gen_loss}
    \end{equation}
\end{enumerate}

\paragraph{Inference.} After each human turn, the context is fed to the RoBERTa classifier. If it predicts "SILENT," the system does nothing. If it predicts "SPEAK," the context is passed to the Llama 3 generator to produce the intervention.

\subsection{Results and Discussion}
We evaluated all baselines on a held-out test set from our generated dataset. The results are summarized in Table \ref{tab:baseline-results}.

\begin{table}[h]
\centering
\caption{Performance of baseline models. The zero-shot model uses the same architecture as the end-to-end model for inference, hence they share the same latency and GPU memory footprint. For the decoupled system, interruption accuracy is from the RoBERTa classifier, while response perplexity is from the Llama 3 generator.}
\label{tab:baseline-results}
\begin{tabular}{@{}lccc@{}}
\toprule
\textbf{Metric} & \textbf{Zero-shot} & \textbf{End-to-end} & \textbf{Decoupled} \\
\midrule
Interruption Accuracy (\%) & 81.72 & 96.59 & 93.18 \\
Response Perplexity       & - & 2.57 & 2.54 \\
\midrule 
Latency (ms/turn)        & \multicolumn{2}{c}{30.12}           & 5.90           \\
GPU Memory (GB)          & \multicolumn{2}{c}{15.47}           & 0.47           \\
\bottomrule
\end{tabular}
\end{table}

Our empirical results show a clear trade-off between decision-making accuracy and computational efficiency. The end-to-end model demonstrates superior performance in the critical task of timing interventions, outperforming the decoupled system in interruption accuracy by over 3 points (96.59\% vs. 93.18\%). In contrast, the zero-shot baseline performs significantly worse at 81.72\%, highlighting the necessity of task-specific fine-tuning. Interestingly, once the decision to intervene is made, the generative quality of the fine-tuned systems is highly comparable, with nearly identical response perplexity.
The main advantage of the decoupled system lies in its significant inference efficiency, making it a highly practical solution for real-world applications. It processes each conversational turn approximately 5 times faster than the end-to-end model (5.90 ms vs. 30.12 ms) while consuming over 30 times less GPU memory (0.47 GB vs. 15.47 GB)\footnote{Computed on NVIDIA RTX 3090 with 24GB GPU memory}. This efficiency comes from the decoupled architecture, which uses a lightweight classifier for the majority of turns and only invokes the resource-intensive LLM when an intervention is necessary. 


%% file: sec/5_conclusion_limitations.tex
\section{Conclusion, Limitations \& Future Work}

In this work, we addressed the "Awareness Gap" inherent in modern Large Language Models, which typically function as reactive agents rather than proactive collaborators. We formalized the "When to Speak" problem and introduced \texttt{DiscussLLM}, a framework and dataset designed to teach LLMs the important skill of timely and valuable intervention in human conversations. Our scalable, two-stage synthetic data generation pipeline successfully produced a large-scale dataset of multi-turn discussions, each containing a natural trigger for a specific AI contribution. By training models to predict a special silent token, we enabled them to actively decide between remaining quiet and offering a helpful response. Our evaluation of two distinct architectures: an integrated end-to-end model and a decoupled classifier-generator system revealed a clear trade-off between intervention accuracy and computational efficiency, providing practical insights for real-world deployment.

This research represents a critical step towards developing always-aware AI agents that can seamlessly integrate into human discussions. The ultimate goal is not to create an AI that constantly interjects, but one that possesses the situational awareness to add value precisely when needed while respecting the natural flow of conversation by remaining silent otherwise. By learning to be a discerning participant, an LLM can evolve from a passive tool into a truly intelligent partner, enhancing collaboration, correcting misinformation, and deepening understanding without being a constant bother.

\subsection{Limitations and Future Work}
Despite the promising results, this work has several limitations that open avenues for future research:

\begin{itemize}
    \item \textbf{Architectural Diversity and Generalization:} Our evaluation is currently confined to the Llama 3 architecture for the generative components. Future work should explore a broader range of LLMs to assess the generalization of our framework and the "When to Speak" skill across different model families. Additionally, the generalization of DiscussLLM to topics beyond Yahoo Answers, or to real-world data with varying discussion styles compared to synthetic generations, is yet to be determined.

    \item \textbf{Fine-Grained and Human-Centric Evaluation:} Our current evaluation is based on proxy metrics such as interruption accuracy and response perplexity. While informative, they do not fully capture the qualitative aspects of a good intervention, such as its helpfulness, relevance, and naturalness. Future work must incorporate more fine-grained metrics and conduct comprehensive human evaluations.

    \item \textbf{Grounded Interventions with External Knowledge:} The current models rely on their internal, parametric knowledge to generate interventions, particularly for \textit{Factual Correction} and \textit{Data Provision}. This can lead to hallucinations or outdated information. A significant next step involves integrating external knowledge sources. Future systems could be enhanced by first predicting the intervention type and then, if necessary, querying a web search engine or a structured database to formulate a more accurate and verifiable response.

    \item \textbf{Data Generation and Grounding:} Although our two-stage synthetic data generation pipeline is highly scalable, the resulting data may not fully capture the complex nuances and unpredictability of real-world human conversations. This initial approach serves to align LLMs for proactivity, but future research could use more grounded data collection methods such as, employing human-in-the-loop systems or large-scale crowd-sourcing to annotate real discussion transcripts.
\end{itemize}